\documentclass[letterpaper, 11pt]{article}

\usepackage[margin=1in]{geometry}
\usepackage{setspace}\onehalfspace
\usepackage{microtype}

\usepackage{amsmath} \allowdisplaybreaks
\usepackage{amssymb}
\usepackage{amsthm} 

\usepackage{natbib}
\usepackage{bibentry}

\usepackage{booktabs} 
\usepackage{graphicx}
\usepackage{appendix}
\usepackage{subfigure}
\usepackage[counterclockwise]{rotating} 
\usepackage[normalem]{ulem} 
\usepackage{showexpl}

\usepackage[hyphens]{url}
\usepackage[bookmarks=false,hidelinks]{hyperref}

\usepackage{tikz}
\usetikzlibrary{automata,calc,trees,positioning,arrows,chains,shapes.geometric,%
decorations.pathreplacing,decorations.pathmorphing,shapes,%
matrix,shapes.symbols,plotmarks,decorations.markings,shadows}

\usepackage{pgf}
\pgfdeclarelayer{background}
\pgfsetlayers{background,main,foreground}

\usepackage{authblk}

\usepackage{hyperref}
\hypersetup{
    bookmarks=true,         
    unicode=false,          
    pdftoolbar=true,        
    pdfmenubar=true,        
    pdffitwindow=false,     
    pdfstartview={FitH},    
    pdftitle={My title},    
    pdfauthor={Author},     
    pdfsubject={Subject},   
    pdfcreator={Creator},   
    pdfproducer={Producer}, 
    pdfkeywords={keyword1, key2, key3}, 
    pdfnewwindow=true,      
    colorlinks=true,       
    linkcolor=blue,          
    citecolor=green,        
    filecolor=magenta,      
    urlcolor=cyan           
}
\usepackage{algorithm}
\usepackage{algpseudocode}

\newcommand{\email}[1]{{\href{mailto:#1}{\nolinkurl{#1}}}}

\usepackage{bm}

\usepackage{xcolor}

\usepackage{etoolbox}
\makeatletter
\def\do#1{\@namedef{#1c}{\ensuremath{\mathcal{#1}}}}
\docsvlist{A,B,C,D,E,F,G,H,I,J,K,L,M,N,O,P,Q,R,S,T,U,V,W,X,Y,Z}
\makeatother

\usepackage{fancyvrb,listings}
\title{Short-term traffic flow forecasting with spatial-temporal correlation in a hybrid deep learning framework}
\author[1]{Wu Yuankai\footnote{Corresponding Author: \email{5433809@bit.edu.cn}}}
\author[2]{Tan Huachun}
\affil[1]{School of Mechanical Engineering, Beijing Institute of Technology}
\affil[2]{School of Mechanical Engineering, Beijing Institute of Technology}

\date{\today}

\begin{document}
\maketitle

\begin{abstract}
Deep learning approaches have reached a celebrity status in artificial intelligence field, its success have mostly relied on Convolutional Networks (CNN) and Recurrent Networks. By exploiting fundamental spatial properties of images and videos, the CNN always achieves dominant performance on visual tasks. And the Recurrent Networks (RNN) especially long short-term memory methods (LSTM) can successfully characterize the temporal correlation, thus exhibits superior capability for time series tasks. Traffic flow data have plentiful characteristics on both time and space domain. However, applications of CNN and LSTM approaches on traffic flow are limited. In this paper, we propose a novel deep architecture combined CNN and LSTM to forecast future traffic flow (CLTFP). An 1-dimension CNN is exploited to capture spatial features of traffic flow, and two LSTMs are utilized to mine the short-term variability and periodicities of traffic flow. Given those meaningful features, the feature-level fusion is performed to achieve short-term traffic flow forecasting. The proposed CLTFP is compared with other popular forecasting methods on an open datasets. Experimental results indicate that the CLTFP has considerable advantages in traffic flow forecasting. in additional, the proposed CLTFP is analyzed from the view of Granger Causality,  and several interesting properties of traffic flow and CLTFP are discovered and discussed
.\\
\noindent\textbf{Traffic flow forecasting, Convolutional neural network, Long short-term memory, Feature-level fusion}
\end{abstract}

\section{Introduction}
The accurate and reliable forecasting of short-term traffic flow is the precursor of a multitude of intelligent transportation systems (ITS) such as proactive dynamic traffic control, intelligent route guidance and intelligent location-based service, thus it is always a hot topic in ITS. Recent developments in information collection and transmission have introduced the notion of big data in the field of ITS \citep{zheng2016big}, which has directed many researches toward data-driven forecasting approaches. 

What are frequently used in data-driven forecasting are the two different approaches which are statistics and neural networks \citep{karlaftis2011statistical}. The statistics such as autoregressive integrated moving average (ARIMA) \citep{min2011real}, Markov chain \citep{qi2014hidden} and Bayesian network \citep{wang2014new}, generally focus on finding the spatial-temporal pattern from a probabilistic perspective and then use that predictive information for forecasting. They can provide some insights on the probabilistic mechanisms generating the traffic data and capture the uncertainty within traffic flow. However, the statistics frequently fail when dealing with nonlinearity within traffic flow because the linear architecture is often used, and they always suffer from curse of dimensionality, which is a common phenomenon in big data era.

Compared with those classical statistical models, the neural networks have several advantages. First, neural networks use tens of thousands of neuron activities to simulate the unknown relationship, and thus they are non-parametric approaches, which are more flexible with input variables. Second, the neural networks are more capable of handling nonlinearity with the help of nonlinear activation functions. Because of those advantages of neural networks and complexity of traffic flow, neural networks are widely used in short-term traffic flow forecasting \citep{chan2012neural}.

Recently, neural networks with deep architectures have proven to be very successful in image, video, audio and language leaning tasks \citep{lecun2015deep}. In short-term traffic forecasting area, though traditionally shallow neural networks are generally adopted, the deep neural networks have also aroused enormous researches' interests. Deep multi-layer fully connected networks are frequently employed in current short-term traffic forecasting \citep{huang2014deep,lv2015traffic}, and pre-training strategies with unsupervised learning algorithms such as Restricted Boltzmann machine (RBM) \citep{hinton2006fast} and Stacked AutoEncoder (SAE) \citep{vincent2008extracting} are often used. Though pre-training strategies significantly promoted the performance of fully-connected networks \citep{hinton2006reducing}, as each neuron in fully-connected layer is connected to every neuron in the previous layer, which makes that fully-connoted networks are expensive in terms of memory and computation. Moreover, there is no assumptions about the features in the fully-connected architecture, thus it is difficult for a fully-connected neural networks to capture representative features from data with plentiful characteristics.  

Also, like frequently studied data in machine learning area such as video and audio, traffic flow data have plentiful characteristics in both time and space domain \citep{7407622,tan2013tensor}. There are some obvious characteristics, for example, in space domain, traffic flow patterns on some location are more likely to have stronger dependencies on nearby locations (topological locality); in time domain, the traffic flow several weeks/days before even has a long-term impact on current traffic flow because of people's traveling habits (long-term memory). A representative characterization of those spatial-temporal features is the key to successful traffic flow forecasting. In recent years, one of the most successful deep neural networks to model topological locality is the convolutional neural network (CNN) \citep{krizhevsky2012imagenet}, it uses filters to find relationships between neighboring inputs, which can make it easier for the network to converge on the correct solution. And one of the most successful architecture to characterize long-term memory is long short-term memory network (LSTM) \citep{graves2013hybrid}, which learns both short-term and long-term memory by enforcing constant error flow through the designed cell state \citep{hochreiter1997long}. 

Motivated by the success of CNN and LSTM, and with consideration of the spatial-temporal characteristics of traffic flow, we propose a novel short-term traffic flow prediction method based on the combination of CNN and LSTM (CLTFP). A deep convolution neural network is utilized to mine the space features of traffic flow data. LSTMs are employed to learn features of both short-term time variation and long-term periodicity. Then we feed those spatial-temporal features into a linear regression layer to predict future traffic flow. In feature-level based data fusion, it is natural to assume that a small portion of features have strong impact \citep{bishop2006pattern}. Thus in order to strengthen the sparsity of features, we add a $l_1$ norm constraint on weights of the regression layer. Finally, we train the neural network end-to-end. Our method is evaluated on traffic flow of a freeway corridor collected from an open datasets, the results show that our method exhibits better performance than state-of-arts. Moreover, we analyze the features captured by CLTFP in terms of incremental predictability, the analysis graphically demonstrate how black-box typed CLTFP understand causality between future-past traffic flow.

\section{Related works}
The deep architecture for short-term traffic flow forecasting has recently been studied in ITS, but mainly on deep fully-connected architecture. \citet{huang2014deep} used a deep belief network to capture the spatial-temporal features of traffic flow and proposed a multi-task learning architecture to perform exit station flow and road flow forecasting. Similarly, \citet{lv2015traffic} proposed a stacked autoencoder model based short-term traffic flow forecasting. \citet{tan2016comparison} investigated the impact of pre-traning with different deep belief networks on the DNN based short-term traffic flow forecasting. \citet{chen2016learning} developed a Stack denoise Autoencoder to learn hierarchical representation of urban traffic flow. \citet{polson2016deep} used a deep neural network architecture to forecast traffic flows during special events. The above approaches have been able to accurately forecast future traffic flow to some degree, but they did not exploit the topological locality and long-term memory of traffic flow, which hindered their predictive power. Motivated by the success of LSTM, \citet{ma2015long} applied LSTM to short-term traffic forecasting, they claimed that LSTM can capture long-term memory of traffic flow. However, despite the efficient usage of long temporal dependency, the spatial dependency is not fully utilized in their work.  

Several attempts have been made to combine CNN and LSTM architectures especially for connection of computer vision and natural language processing. For example, several methods have made use of CNN features and LSTM for image/video description generation \citep{vinyals2015show,yao2015describing,peris2016video}. The combination of CNN and LSTM have also been successfully applied to visual activity recognition \citep{donahue2015long}, sentiment analysis \citep{wang2016dimensional} and video classification \citep{wu2015modeling}. As traffic flow data share some common properties with visual data \citep{yang2014detecting} and language, the success of combining CNN and LSTM on computer vision and language processing indicates the potential of such combination on traffic flow forecasting. 

\section{Model}
\begin{figure}[h]
\begin{center}
\includegraphics[width=0.45\textwidth]{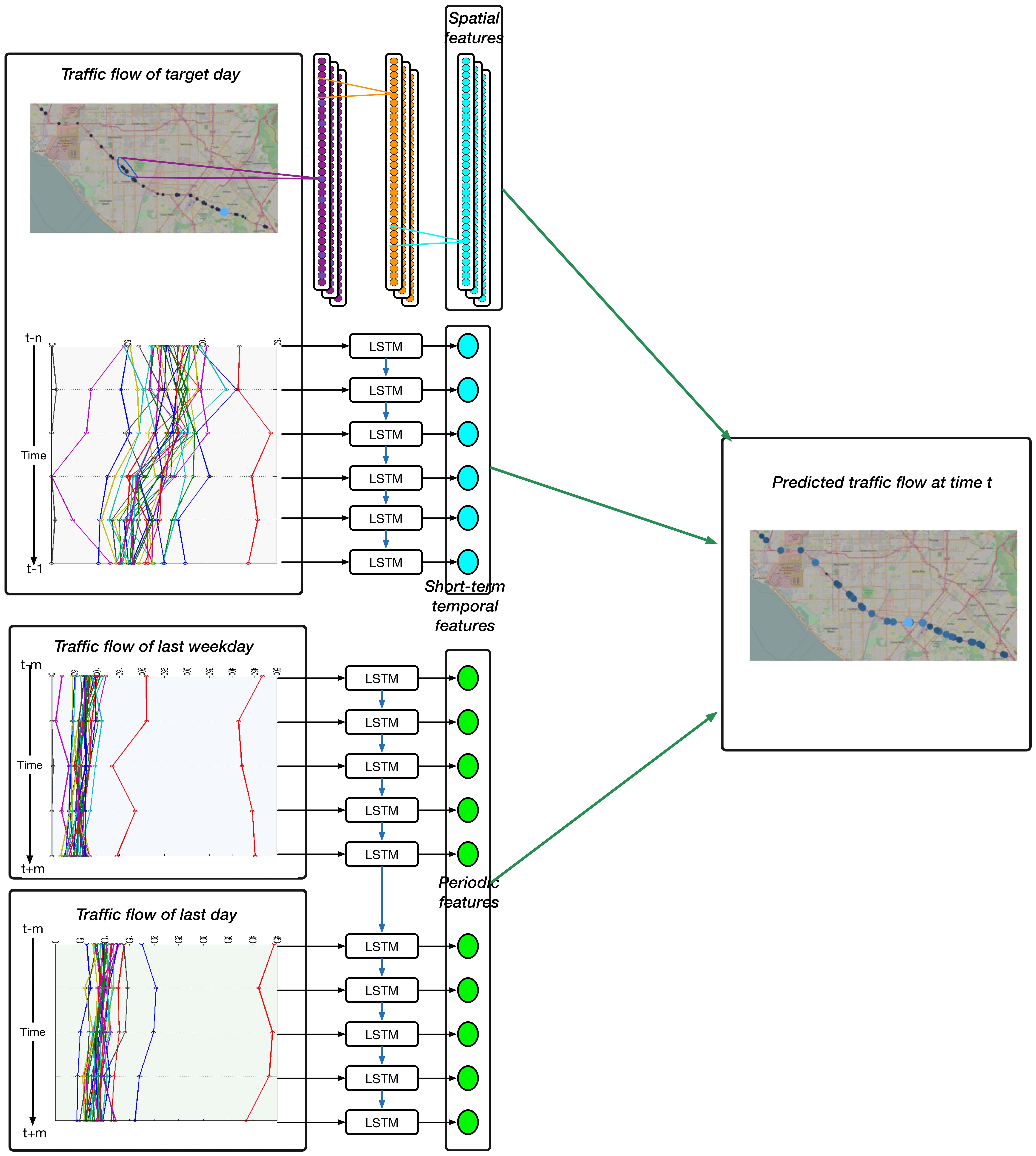}
\caption{CLTFP, our model, is based end-to-end on a neural network consisting of a 1D CNN (capture spatial features), two LSTM RNNs (capture short-term and periodic features) and followed by a fully connected layer to fuse all features to forecast traffic flow at target time point $t$.} 
\label{Fig1}
\end{center}
\end{figure}

In this section, we describe our CNN-LSTM based short-term traffic flow prediction method (CLTFP). See Fig. 1 for the graphical illustration of the proposed model. It can be found from Fig.\ref{Fig1} that CLTFP consists of a 1D CNN, two LSTM RNNs and a fully-connected layer, we will describe each part in the following subsection. 

\subsection{Spatial features captured by CNN}
Suppose we need to forecast traffic flow of $p$ locations $\{s_i\}{^p}{_{i=1}}$ in $(t,t+1,\cdots,t+h)$, in which $h$ is the prediction horizon. The historical traffic flow data of $p$ locations $\{s_i\}{^p}{_{i=1}}$ in $(t-n,t-1)$ are traditionally used as inputs for generating prediction in $(t,t+1,\cdots,t+h)$. If we put the historical data together, we can get a data matrix 
\begin{equation}
\mathbf{S} = \begin{bmatrix}  
      S_1  \\[0.2em]
       S_2            \\[0.1em]
       .                \\[0.1em]
       .                \\[0.1em]
       .                \\[0.1em]
       S_p
     \end{bmatrix}    =     \begin{bmatrix}  
      s_1(t-n) & s_1(t-n+1) & \cdots & s_1(t-1)  \\[0.1em]
      s_2(t-n) & s_2(t-n+1) & \cdots & s_2(t-1)   \\[0.1em]
       .       &    .  &   &.\\[0.1em]
       .        &    .   & &.\\[0.1em]
       .         &    .   & &.\\[0.1em]
        s_p(t-n) & s_p(t-n+1) & \cdots & s_p(t-1)
     \end{bmatrix}.
\label{input_matrix}
\end{equation}

The traffic flow usually depends on traffic flow of that location and its neighbors. As Convolutional Neural Networks (CNNs) have been successful in handling data representation with a locality structure, we naturally adapt a 1-dimensional CNN to capture spatial features of traffic flow. Our 1D CNN does not attack time mode. Instead, the time dimension is treated as channels of an image, which means that we only perform convolution on vectors $T_q= \begin{bmatrix} s_1(t-n+q), s_2(t-n+q), \cdots, s_p(t-n+q)  \end{bmatrix}^T$($0 \geq q \leq n-1$) of matrix $\mathbf{S}$ in Eq.(\ref{input_matrix}). 

For locations of a freeway corridor given in Fig.\ref{Fig1}, the traffic travel through from upstream $s_1$ to downstream $s_p$, the conventional 1D CNN is naturally exploited to capture spatial features of such a transportation network, where the $k$-th feature map is obtained as follows
\begin{equation}
h^k_q = o_c(w^k_q \ast T^k_q + b^k_q),
\end{equation}
where $w^k_q$ is the weights vector, $b^k_q$ is the bias, $o_c$ denotes a nonlinear activation and $\ast$ denotes the convolution. The pooling layers are not employed in our model. Because it is evident that an all convolution net achieves better performance on small images recognition \citep{springenberg2014striving}, and the space dimension of traffic data on short-term traffic forecasting task is always limited.

There are many more complex transportation networks than the one given in Fig.\ref{Fig1}, e.g. a transportation network of a big city. In these cases, the conventional 1D CNN can not be used without any modifications. However, the traffic flow in any transportation network always has some graph structures \citep{shahsavari2015short}, Thus the CNN on graph-structured data proposed by \citet{henaff2015deep} can be an alternative for such transportation networks.

\subsection{Short-term temporal features}
As stated in \citet{ma2015long}, the traditional forecasting models mainly suffer from two drawbacks in handling time mode information of traffic flow: (1) Traditional methods especially traditional RNNs are difficult to train if the traffic flow series has long time lags, which means traditional RNNs will provide poor performance if the time window size $n$ in Eq.(\ref{input_matrix}) is too large. (2) It is difficult to find the optimal time window size $n$, as the correlation between traffic flow of different time points is affected by many complex factors such as weather, speed and unpredictable incident. The LSTM is one of the more practical ways to tackle these issues, thus we propose to use LSTM to capture the time mode information of traffic flow. Different from the work of \citet{ma2015long}, which directly used LSTM to generate predictions of traffic flow, we use LSTM to generate time features at each time point and build a deeper and more complex traffic flow forecasting model.

\begin{figure}[h]
\begin{center}
\includegraphics[width=0.37\textwidth]{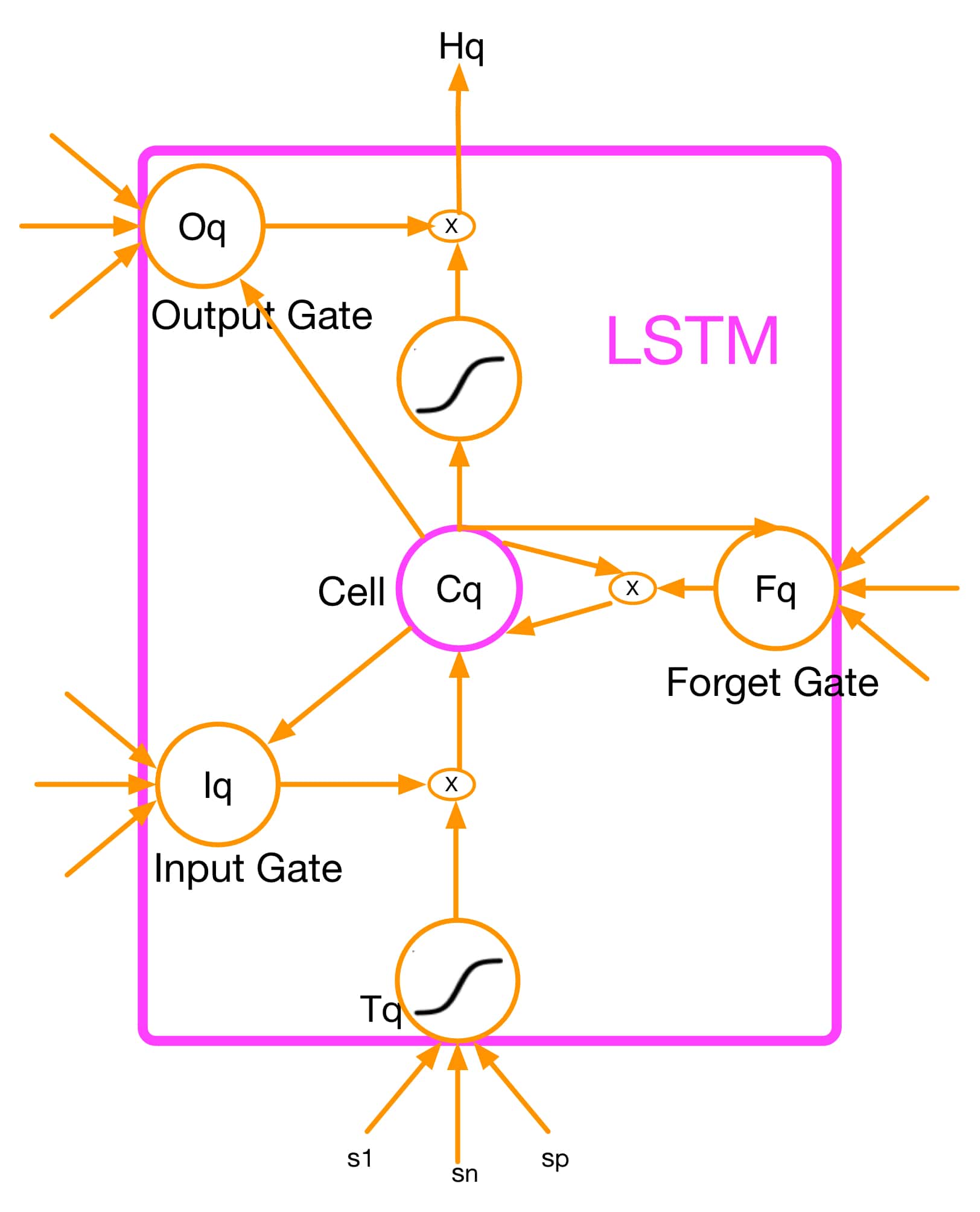}
\caption{LSTM structure}
\label{Fig2}
\end{center}
\end{figure}

Similar to the traditional RNNs, a LSTM structure is composed of one input layer, one or several hidden layers and one output layer. The core idea of LSTM is the memory cell in hidden layers, which is designed to avoid the gradient vanish and explosion in traditional RNNs. As shown in Fig. \ref{Fig2}, a memory cell contains four main parts: an input gate, a neuron with a self-recurrent connection, a forget gate and an output gate.
 
For the generation of short-term temporal features, the inputs of LSTM is denoted as $T=(T_0,T_1,\cdots,T_{n-1})$ where $T_q= \begin{bmatrix} s_1(t-n+q), s_2(t-n+q), \cdots, s_p(t-n+q)  \end{bmatrix}^T$ in Eq (\ref{input_matrix}), and the output temporal features in each historical time point is denoted as $H=(H_0,H_1,\cdots,H_{n-1})$, $n$ is the time window size. The generated temporal features are iteratively calculated by following equations:
\begin{equation}
I_q = \sigma(W_i T_q+ U_i H_{q-1}+W_{ci} \cdot C_{q-1}+b_i),
\end{equation}
\begin{equation}
F_q = \sigma(W_f T_q+ U_f H_{q-1}+W_{cf} \cdot C_{q-1}+b_f),
\end{equation}
\begin{equation}
C_q = I_q \cdot \sigma{_h}(W_c T_q+U_c H_{q-1}+b_c)+F_q \cdot C_{q-1} ,
\end{equation}
\begin{equation}
O_q = \sigma(W_o T_q+ U_o H_{q-1}+V_o \cdot C_q+b_q),
\end{equation}
\begin{equation}
H_q = O_q \cdot \sigma{_h}(C_q).
\end{equation}
where $\cdot$ represents the Hadamard product of two vectors, and $\sigma(.)$ and $\sigma{_h}(.)$ are activation functions. $\sigma(.)$ is traditionally set to be a function of domain $[0,1]$ to control the information flow through time. And $\sigma{_h}(.)$ is often set to be a centered activation function. 

\subsection{Periodic features}
People are used to repeating some behaviors on a same time period of day, e,g. people routinely go to work in the morning and go home in the evening, This is why we can observe a strong periodicity within traffic flow. The periodicity of traffic flow have been identified as a major contributing factor for the traffic flow forecasting. A desirable model that successfully characterize the periodicities can accurately forecast future traffic flow. In this subsection, we propose to use LSTM to capture the features of periodicities for traffic flow forecasting. The inputs for periodicities at time point $t$ are given as following:
\begin{equation}
\mathbf{S^d} =      \begin{bmatrix}  
      s_1(t^d-n^d) & s_1(t^d-n^d+1) & \cdots & s_1(t^d+n^d)  \\[0.1em]
      s_2(t^d-n^d) & s_2(t^d-n^d+1) & \cdots & s_2(t^d+n^d)   \\[0.1em]
       .       &    .  &   &.\\[0.1em]
       .        &    .   & &.\\[0.1em]
       .         &    .   & &.\\[0.1em]
        s_p(t^d-n^d) & s_p(t^d-n^d+1) & \cdots & s_p(t^d+n^d),
     \end{bmatrix}.
   \label{daily}
\end{equation}
\begin{equation}
\mathbf{S^w} =      \begin{bmatrix}  
      s_1(t^w-n^w) & s_1(t^w-n^w+1) & \cdots & s_1(t^w+n^w)  \\[0.1em]
      s_2(t^w-n^w) & s_2(t^w-n^w+1) & \cdots & s_2(t^w+n^w)   \\[0.1em]
       .       &    .  &   &.\\[0.1em]
       .        &    .   & &.\\[0.1em]
       .         &    .   & &.\\[0.1em]
        s_p(t^w-n^w) & s_p(t^w-n^w+1) & \cdots & s_p(t^w+n^w),
     \end{bmatrix}.
\end{equation}
where $t^d$ and $t^w$ denote the same time point of $t$ in last day and last weekday respectively, $n^d$ and $n^w$ denote the time lags of daily periodicity and weekly periodicity respectively. 

For features of daily periodicity, the input of LSTM is denoted as $T^d=(T^d_0,T^d_1,\cdots,T^d_{2n^d})$ where $T^d_q= \begin{bmatrix} s_1(t^d-n^d+q), s_2(t^d-n^d+q), \cdots, s_p(t^d-n^d+q)  \end{bmatrix}^T$ in Eq (\ref{daily}). The same input can be easily adapted to weekly periodicity. It is obvious that there exists correlation between $\mathbf{S^d}$ and $\mathbf{S^d}$, so as illustrated in Fig.\ref{Fig1}, the connection between LSTMs for daily periodicity and weekly periodicity are added. With the LSTM architecture given in Fig.\ref{Fig1}, we can capture features $H^w_0,H^w_1,\cdots,H^w_{2n^w}$ of weekly periodicity and features $H^d_0,H^d_1,\cdots,H^d_{2n^d}$ of daily periodicity to forecast future traffic flow at time point $t$.  

\subsection{Feature-level fusion}
As shown in Fig. \ref{Fig1}, the proposed CLTFP captures spatial features $h^1,h^2,\cdots,h^{n_c}$, short-term temporal features $H_0,H_1,\cdots,H_{n-1}$, weekly periodic features $H^w_0,H^w_1,\cdots,H^w_{2n^w}$, and daily periodic features $H^d_0,H^d_1,\cdots,H^d_{2n^d}$. In order to fuse them to perform short-term traffic flow forecasting, we concatenate all the features sequentially into a feature vector, then add a regression layer to perform forecasting. The objective function of regression is the sum of square errors between predicted value $s^{\epsilon}_1(t), s^{\epsilon}_2(t), \cdots, s^{\epsilon}_n(t)$ and future value $s_1(t), s_2(t), \cdots, s_n(t)$. There may be redundancies between the features captured by CLTFP. To handle the feature redundancy problem, we add a sparsity regularization in weights of fully connection layer thus our model is likely to assign a weight close to zero to redundant feature. 

\section{Experiments}
In this section, we use traffic flow data from PeMS(http://pems.eecs.berkeley.edu/)  to evaluate the proposed CLTFM. The CLTFM is compared with several state-of-the-art forecasting methods with deep architecture: LSTM \citep{ma2015long}, SAE \citep{lv2015traffic}, a shallow neural network and the gradient boosting regression tree (GBRT) method \citep{zhang2015gradient}. In additional, we perform a study to discover the performance contribution of different features of CLTFM from the view of Granger Causality.  All experiments are performed by a PC (CPU: Intel Xeon(R) E5-2620 2.1GHz, 64GB memory, GPU: NVIDIA® Tesla K40C).

\subsection{Datasets}
\begin{figure}[h]
\begin{center}
\includegraphics[width=0.4\textwidth]{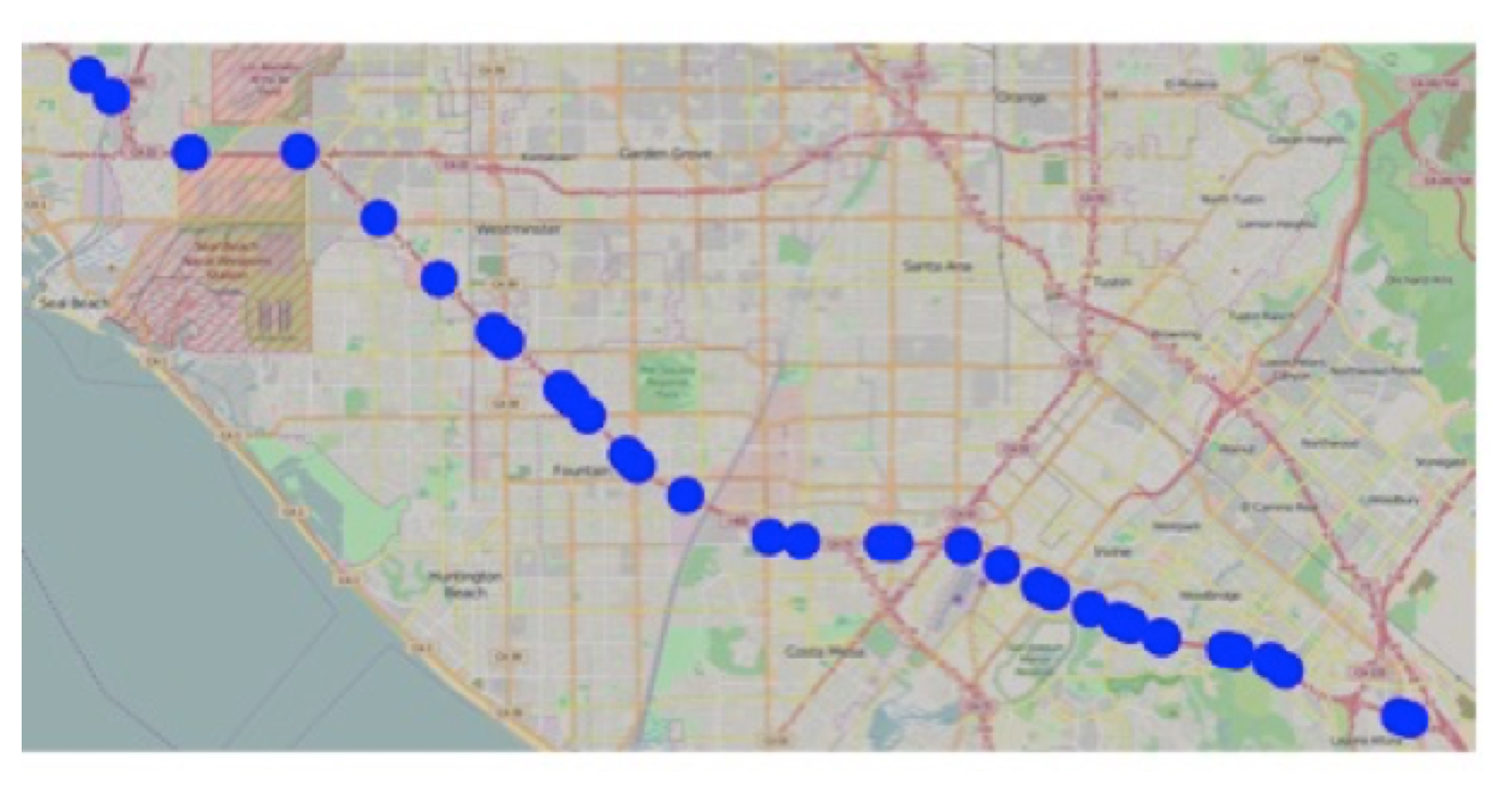}
\caption{Traffic flow locations studied in this paper}
\label{Fig3}
\end{center}
\end{figure}

The peculiar traffic flow data from PeMS throughout North-bound I-405 trip are used for our experiments. Traffic flow of 33 locations given in Fig.\ref{Fig3} on this trip are used for our study. The particular time period used in this paper is from 01/04/2014 to 30/06/2015. The traffic volume are aggregated every 5 min. Thus, one detector preserves 288 data points per day. We use earlier 110000 past-future pairs to train all models, and the rest pairs are used as test data.

\subsection{Experimental setup} 
For all methods, time window size $n$ of $\mathbf{S}$ is set as $15$, which means that 75 min historical data are used to perform forecasting of next 5 min. The time lags of daily periodicity $n^d$ and weekly periodicity $n^w$ for long-term inputs of CLTFP are set as $6$, which means that the traffic flow before and after 30 minutes in previous day and weekday are used to generate forecasting.

For 1D CNN structure of CLTFP, a 3-layer fully convolution structure is used, there are 30 filters in each layer, the filter lengths of first 2 layers are set as 3, the filter length of last layer is set as 2, SReLu \citep{jin2015deep} is used as the activation function of CNN. For the LSTM capturing short-term temporal features $H_0,H_1,\cdots,H_{n-1}$, the feature dimension of each time point is set as 40. For the LSTM capturing long-term features $H^d_0,H^d_1,\cdots,H^d_{2n^d}$ and $H^w_0,H^w_1,\cdots,H^w_{2n^w}$, the feature dimension of each time point is set as 25. For the regression layer of CLTFP, the $l_1$ norm regularizer on weights is set as 0.002.

CLTFP are trained based on Adamax optimizer \citep{kingma2014adam}, we randomly select 10\% training data as validation dataset to control earlystopping. The architecture of CLTFP are built upon Keras framework \citep{chollet2015}. The structures and parameters for other methods are set according to the reports on corresponding papers. 

\subsection{Comparison results}
In this paper, the mean absolute percentage error ($MAPE$) is used to compare the performance of traffic forecasting. The $MAPE$ will be lower if the traffic volumes are higher. In observance of this, this paper also applies the mean absolute error ($MAE$) as a complementary measure for $MAPE$,
\begin{equation}
MAE=\frac{1}{n_p}\sum_{t=1,s=1}^n|z_{st}-N_{st}|,
\end{equation}
\begin{equation}
MAPE=\frac{1}{n_p}\sum_{t=1,s=1}^n\frac{|z_{st}-N_{st}|}{N_{st}}\times100\%,
\end{equation}
where $z_{st} =$ predicted traffic flow at time point $t$ on location $s$; $N_{st} =$ actual traffic flow; $n_p =$ number of predictions. The aim of indexes $MAE$ and $MAPE$ is to measure the errors between predicted values and actual values. The forecasting correctness of spatial distribution is also an important index for this comparison as we perform prediction on multiple locations, thus we define an average correlation error (ACE) to measure the ability of spatial distribution forecasting:
\begin{equation}
ACE = \frac{1}{n_t} \sum_{t=1}^n{Corr(z_{:t},N_{:t})},
\end{equation}
where $z_{:t}=$ predicted traffic flow vector at time point $t$; $N_{:t} =$ actual traffic flow vector; $n_t=$ number of prediction steps. 

Table.\ref{table1} gives quantitative results of CLTFP, LSTM, SAE, shallow NN and GBRT, it can be found that CLTFP achieves better performance than other methods in terms of prediction accuracy and spatial distribution. The reason is that CLTFP makes full use of spatial distribution, short-term temporal variability and long-term periodicities.
\begin{table}[!t]
\renewcommand{\arraystretch}{1.3}
\caption{The quantitative results of different methods}
\label{table1}
\centering
\begin{tabular}{ c c c c}
\hline
\hline
 features & MAE & MAPE & ACE\\
\hline
 CLTFP & 19.37 &7.36\% &0.9263 \\
 \hline
 LSTM &21.53 &8.55\% &0.9137 \\
 \hline
 SAE &20.36 &8.07\% &0.9198 \\
 \hline
 NN &20.61 &8.31\% &0.9174 \\
 \hline
 GBRT &22.52 & 8.52\% &0.9109\\
 \hline
\hline
\end{tabular}
\end{table}

\subsection{Analysis of Features}
One constant criticism of using neural networks on transportation area has been that they are black box models, with little understanding of how the networks work and what knowledge the networks find from data. Recently, it is found that Granger Causality, which characterizes the causality based on incremental predictability, can be adopted to understand black-box typed prediction approaches \citep{li2015robust}. In this subsection, we focus on analyzing proposed CLTFP from the view of Granger Causality.

\begin{figure*}[ht]
  \centering
  \subfigure[prediction by LASSO using S]{
    \label{fig:subfig:a} 
    \includegraphics[width=0.2\textwidth]{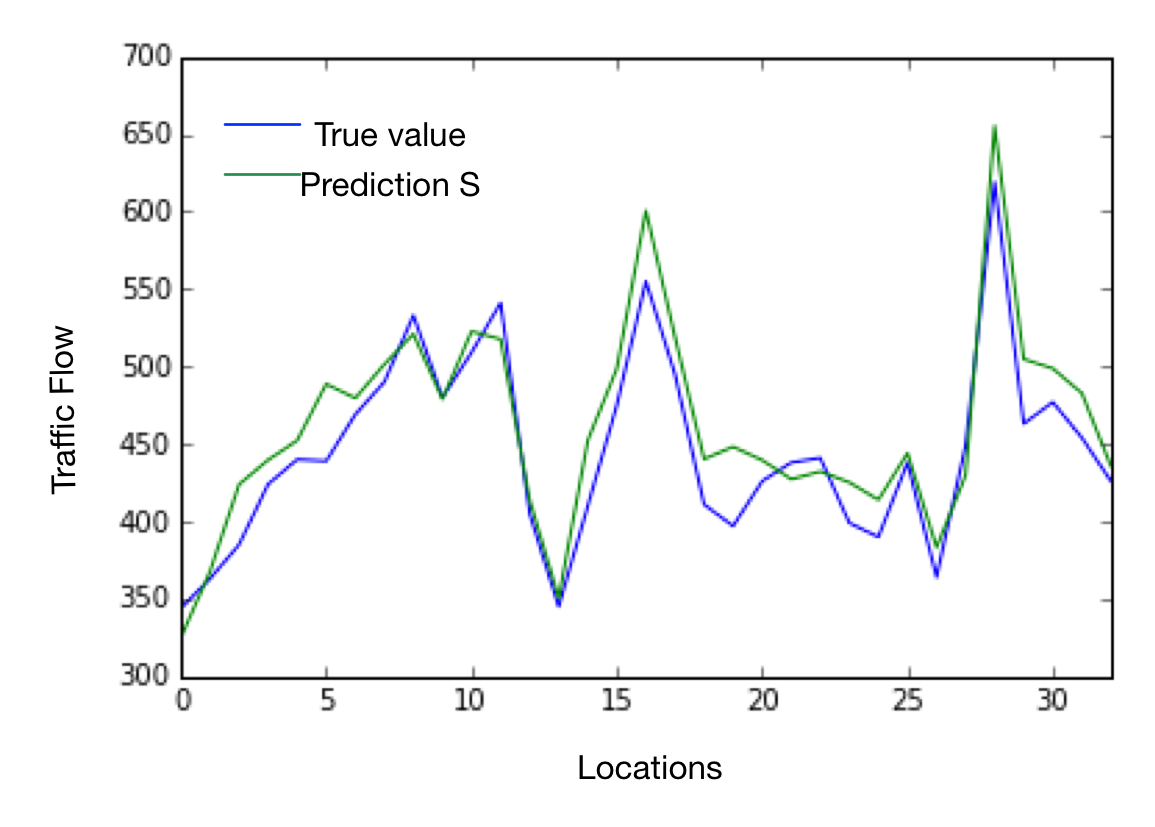}}
  \hspace{0.0in}
  \subfigure[prediction by LASSO using T]{
    \label{fig:subfig:b} 
    \includegraphics[width=0.2\textwidth]{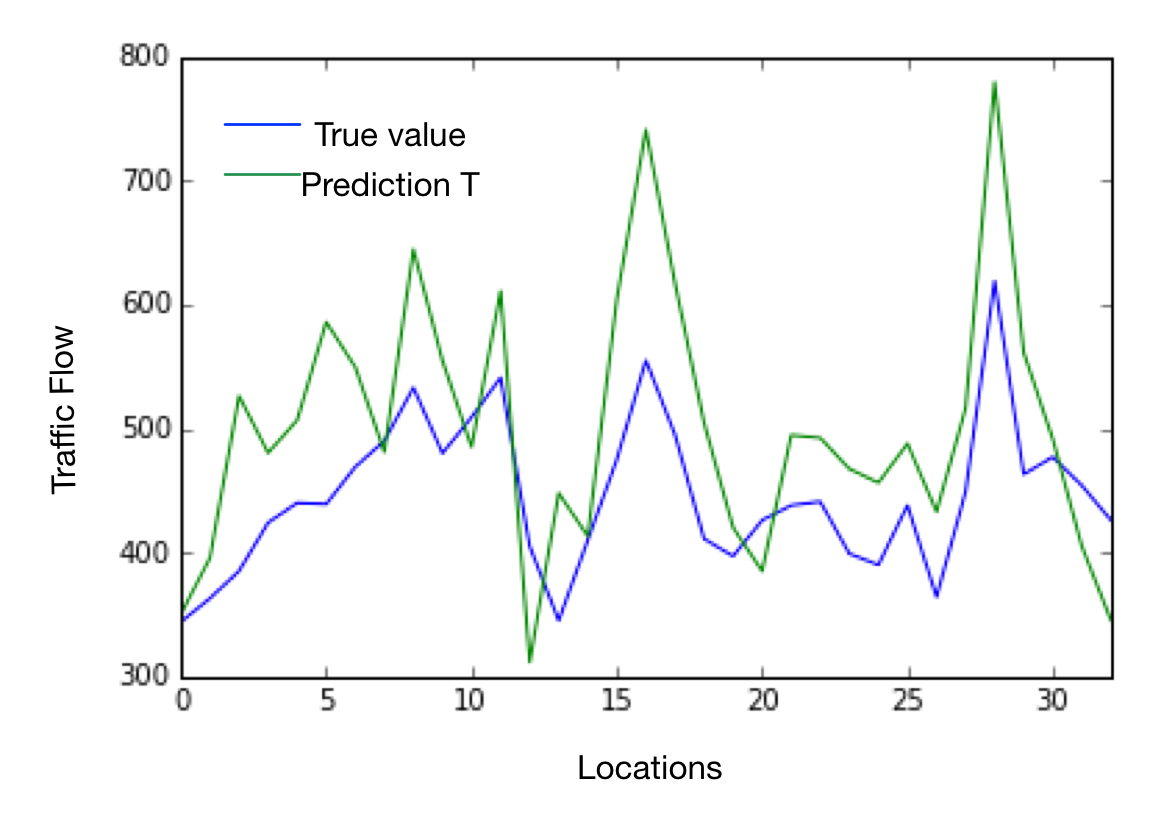}}
  \hspace{0.0in}
    \subfigure[prediction by LASSO using P]{
    \label{fig:subfig:a} 
    \includegraphics[width=0.2\textwidth]{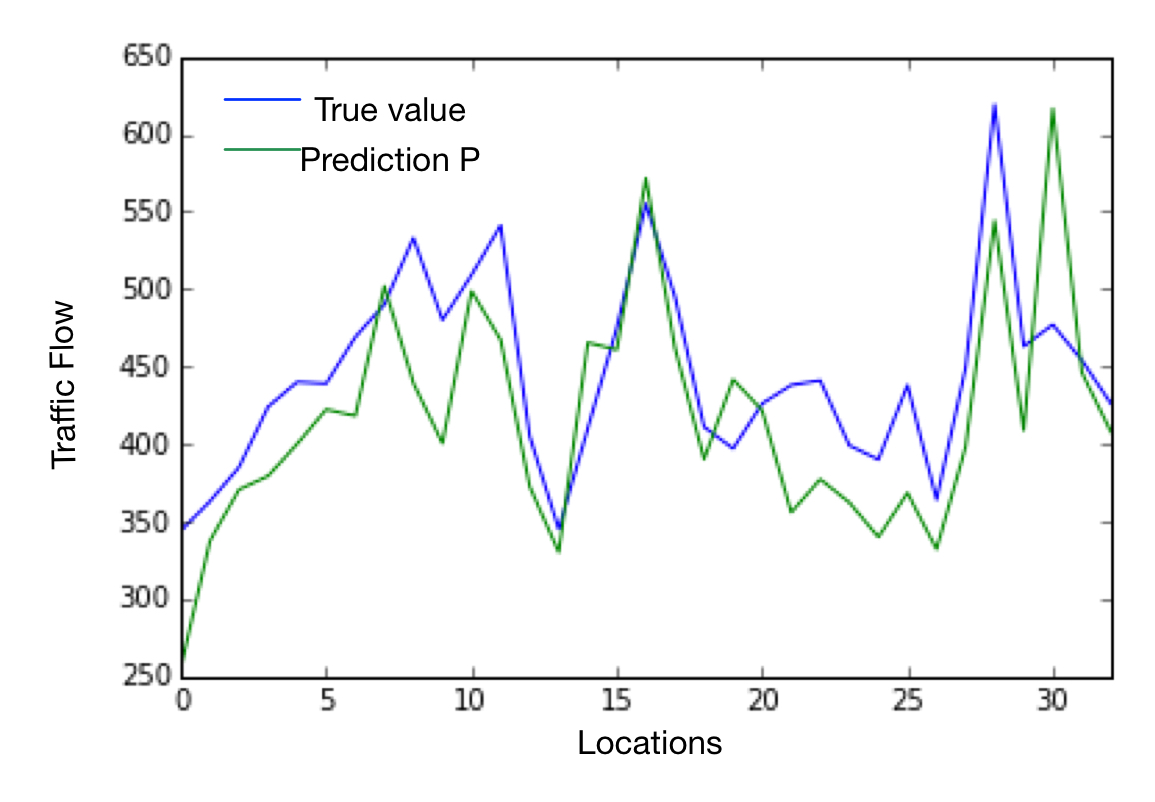}}
  \hspace{0.0in}
   \subfigure[prediction by LASSO using P+T]{
    \label{fig:subfig:a} 
    \includegraphics[width=0.2\textwidth]{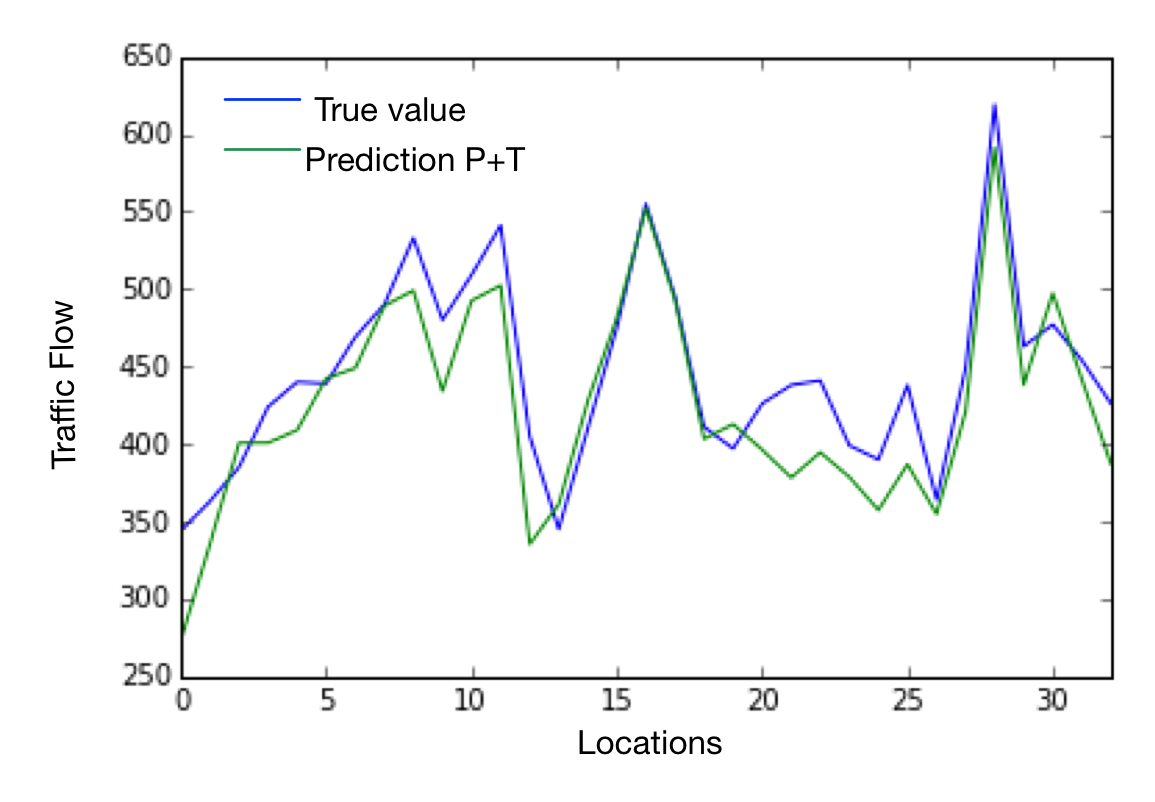}}
  \hspace{0.0in}
   \subfigure[prediction by LASSO using T+S]{
    \label{fig:subfig:a} 
    \includegraphics[width=0.2\textwidth]{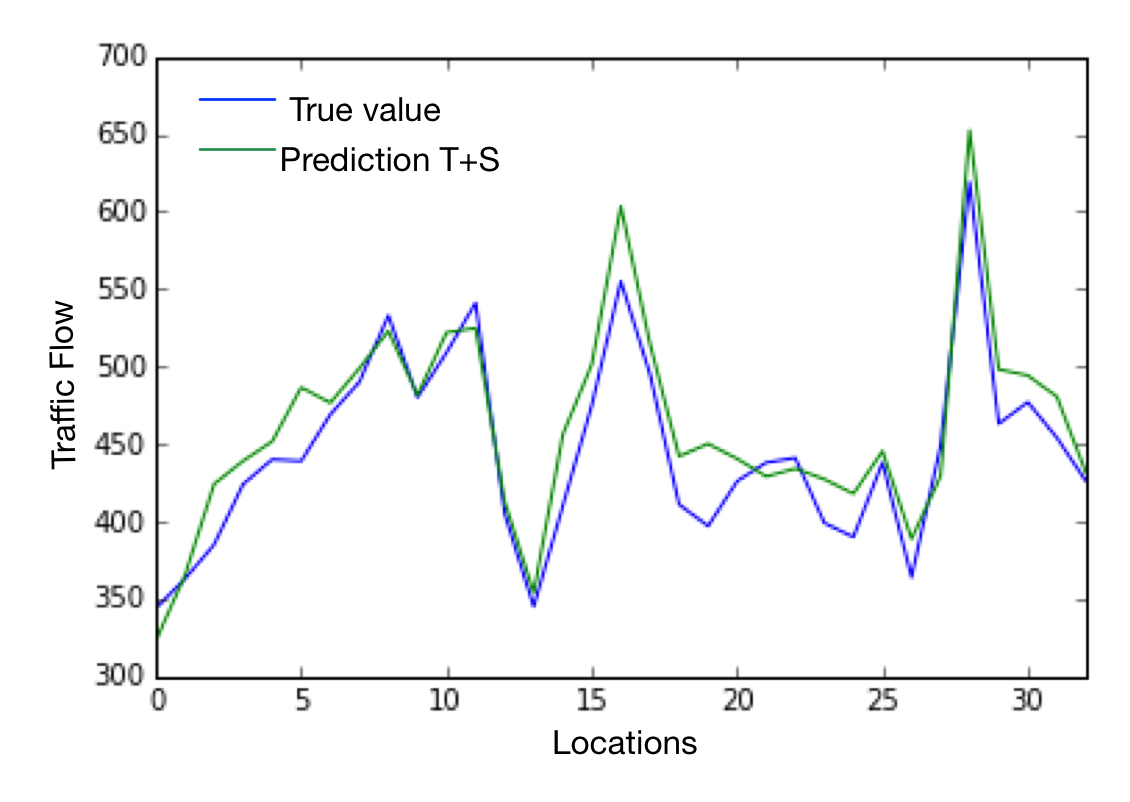}}
  \hspace{0.0in}
  \subfigure[prediction by LASSO using P+S]{
    \label{fig:subfig:a} 
    \includegraphics[width=0.2\textwidth]{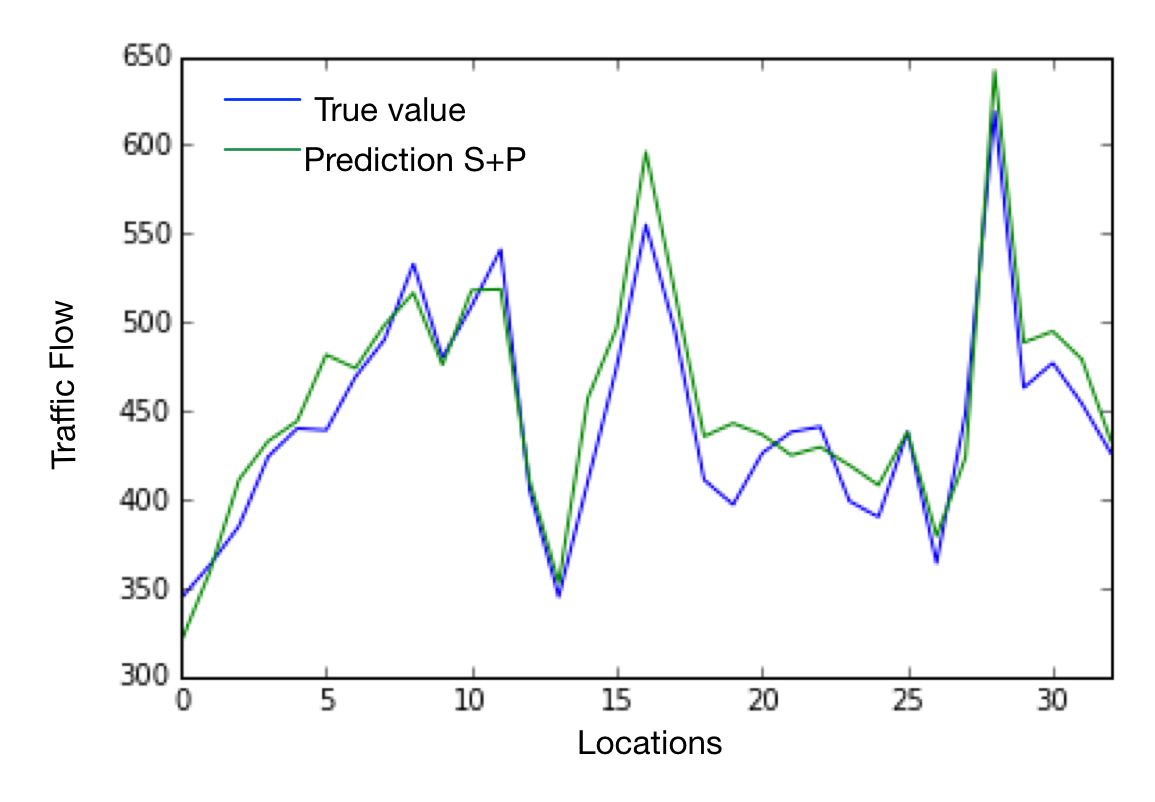}}
     \hspace{0.0in}
  \subfigure[prediction by LASSO using P+T+S]{
    \label{fig:subfig:a} 
    \includegraphics[width=0.2\textwidth]{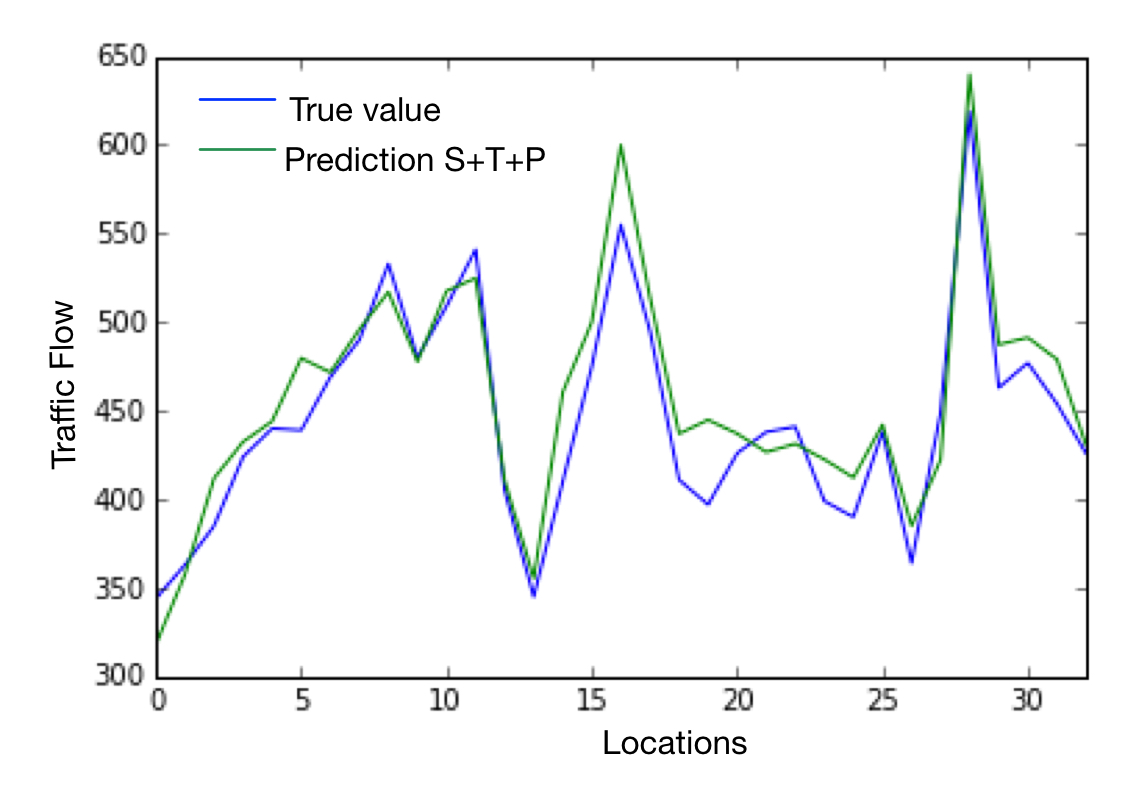}}
     \hspace{0.0in}
  \subfigure[prediction by CLTFP]{
    \label{fig:subfig:a} 
    \includegraphics[width=0.2\textwidth]{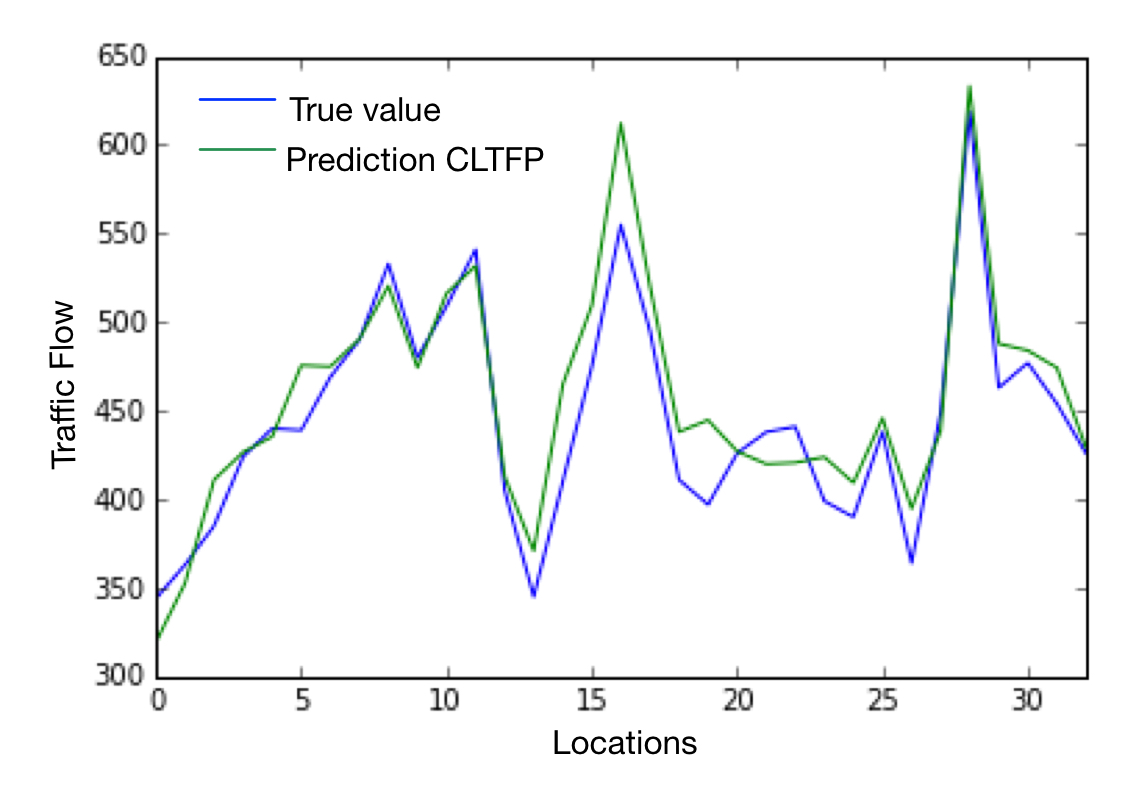}}
  \caption{Forecasting results of different LASSO models and proposed CLTFP at one time point}
  \label{quare} 
\end{figure*}

The experimental analysis is conducted as following: we leverage available spatial features (S), short-term temporal features (T) and periodic features (P) generated from well-trained model CLTFP to train several Lasso models, and then conduct a comparison between forecasting results of those Lasso models with different combinations of features. All Lasso models are fit with Least Angle Regression, the penalty terms of $l_1$ priors are all set as $0.002$.

The experimental results are given in Table. \ref{table2}. This table is quite revealing in several ways. 1. We can generally conclude that more types of feature help to build better prediction results. As the improved predictability can be achieved by adding those features, we can draw that future traffic flow are dependent on all those information from the view of Granger Causality. 2. The model with feature S significantly outperforms the model with feature T and P, it suggests that future traffic flow of the studied corridor is heavily dependent on spatial information of near-term traffic flow though the temporal information in both near-term and long-term (last week and last day) have some influence. 3. The model with feature P achieves lower errors than model with feature T, however, it has weaker predictability on spatial distribution of future traffic flow. It suggests that the travel habits have more influence on the total traffic flow, but near-term traffic flow on transportation network is more related to the future distribution of traffic flow. 4. The model with features S, T and P even outperforms our well-trained CLTFP in terms of $MAE$, $MAPE$ and $ACE$. It indicates that the performance of traffic flow forecasting can be promoted by using a proper regression model on features generated from a well-trained neural networks. The similar phenomenons can be also found in Fig. \ref{quare}, which gives quatitative visualization of forecasting results at one time point.

\begin{table}[!t]
\renewcommand{\arraystretch}{1.3}
\caption{The quantitative results of different features and their combinations (S: spatial features, T: Short-term temporal features, P: Periodicity features)}
\label{table2}
\centering
\begin{tabular}{ c c c c}
\hline
\hline
 methods & MAE & MAPE & ACE\\
\hline
S & 19.83 &7.91\% &0.9302 \\
 \hline
T & 54.43 &28.40\% &0.8532 \\
 \hline
P &40.79 &18.06\% &0.7996 \\
\hline
P+T &35.07 &14.59\% &0.8763\\
\hline
S+T &19.61 &7.49\%  &0.9314\\
\hline 
S+P &19.47 &7.38\%  &0.9312\\
\hline 
S+T+P &19.32 &7.29\% & 0.9323\\
 \hline
\hline
\end{tabular}
\end{table}

\section{Conclusions and future work}
A novel deep learning based short-term traffic flow forecasting method CLTFP combined with CNN and LSTM is proposed in this paper, the forecasting results of CLTFP are encouraging, it indicates the potential of CNN and LSTM on transportation applications. Moreover, incremental predictability is applied to analyze the black-box typed forecasting method, the analysis shows that neural network based forecasting method can provide many meaningful knowledges of traffic flow. 

This proposed CLTFP admits many improvements and extensions:

\noindent 1. The features captured by LSTM achieves only modest forecasting accuracy, some more complex structures, for example, convolutional LSTM structure \citep{xingjian2015convolutional} can be an alternative. 

\noindent 2. Traffic flow are affected by many other factors such as weather, social event and state of the roads. How to exploit those information as auxiliary information is a future direction.

\noindent 3. The applications of our model on general transportation network and similar spatial-temporal data on other domain are straightforward extensions.

\bibliographystyle{AAAI}		
\bibliography{wu}			

\end{document}